\documentclass{isprs} 
\usepackage{subfigure}
\usepackage{setspace}
\usepackage{geometry} 
\usepackage{epstopdf}
\usepackage[labelsep=period]{caption}  
\usepackage[british]{babel} 
\usepackage[hang]{footmisc}
\usepackage{hyperref}
\usepackage{amsfonts}
\usepackage{amsmath}
\usepackage{amssymb}
\usepackage{booktabs}
\usepackage{xcolor}
\usepackage{mathtools}
\usepackage{gensymb}
\usepackage[ruled,vlined]{algorithm2e}
\SetKwProg{Init}{initialize}{}{}

\DeclarePairedDelimiter\abs{\lvert}{\rvert}%
\makeatletter
\let\oldabs\abs
\def\abs{\@ifstar{\oldabs}{\oldabs*}}
\begin{document}

\title{GRAPH-THEORETICAL APPROACH TO ROBUST 3D NORMAL EXTRACTION OF LIDAR DATA}
\date{}

\author{Arpan Kusari$^{* \dagger}$, Wenbo Sun$^\dagger$}

\address{University of Michigan Transportation Research Institute, University of Michigan - \{kusari, sunwbgt\}@umich.edu}


\commission{II}{YY} 
\workinggroup{3} 
\icwg{}   

\abstract{
Low dimensional primitive feature extraction from LiDAR point clouds (such as planes) forms the basis of majority of LiDAR data processing tasks. 
A major challenge in LiDAR data analysis arises from the irregular nature of LiDAR data that forces practitioners to either regularize the data using some form of gridding or utilize a triangular mesh such as triangulated irregular network (TIN). While there have been a handful applications using LiDAR data as a connected graph, a principled treatment of utilizing graph-theoretical approach for LiDAR data modelling is still lacking.
In this paper, we try to bridge this gap by utilizing graphical approach for normal estimation from LiDAR point clouds.
We formulate the normal estimation problem in an optimization framework, where we find the corresponding normal vector for each LiDAR point by utilizing its nearest neighbors and simultaneously enforcing a graph smoothness assumption based on point samples. This is a non-linear constrained convex optimization problem which can then be solved using projected conjugate gradient descent to yield an unique solution. As an enhancement to our optimization problem, we also provide different weighted solutions based on the dot product of the normals and Euclidean distance between the points. 
In order to assess the performance of our proposed normal extraction method and weighting strategies, we first provide a detailed analysis on repeated randomly generated datasets with four different noise levels and four different tuning parameters. Finally, we benchmark our proposed method against existing state-of-the-art approaches on a large scale synthetic plane extraction dataset. The code for the proposed approach along with the simulations and benchmarking is available at \url{https://github.com/arpan-kusari/graph-plane-extraction-simulation}.}

\keywords{Normal estimation, Graphical approach, Optimization, Projected Conjugate Gradient Descent}

\maketitle
\def\thefootnote{$*$}\footnotetext{Corresponding author}\def\thefootnote{\arabic{footnote}}
\def\thefootnote{$\dagger$}\footnotetext{These authors contributed equally to this work}\def\thefootnote{\arabic{footnote}}

\section{INTRODUCTION}
\label{sec:intro}

There have been a proliferation of laser scanners and three dimensional (3D) cameras in domains as varied as mobile robotics \cite{suger2015traversability}, autonomous vehicles \cite{urmson2008autonomous} and advanced manufacturing \cite{hitachi} along with traditional domains such as land surveying. These 3D scanners measure 3D locations of points on objects and store them as point cloud data. Analyzing the point cloud data is a fundamental task in mapping and navigation applications. For example, urban and indoor environments usually comprise of a large amount of planar surfaces. The planar information can be collected as 3D point clouds via light detection and ranging (LiDAR) techniques. Identifying planes from 3D point clouds is a non-trivial task in the presence of multiple inlier structures and contamination of observed data with noise \cite{amayo2018geometric}.

Generally speaking, there are two disparate domains which utilize 3D point clouds and perform plane identification. While Photogrammetry utilizes plane fitting methods such as region growing \cite{xu2017segmentation}, 3D Hough transform \cite{duda1972use} and RANSAC \cite{fischler1981random} to cluster points belonging to an actual planar surface, Computer Vision typically relies on per point normal estimation that is computed based on neighboring points on the local planar neighborhoods \cite{jordan2014quantitative}. In this research, we take the view that computing the per-point normal is a lower level function which can then be clustered to derive the planar surfaces as detailed in \cite{hoover1996experimental}. 

Given that 3D point cloud data is inherently unstructured, our proposed approach is established based on graphs which can provide an inherently principled way of connecting adjacent points in a global manner and promote relationship between them based on the given edge characteristics. Although the graphical formulation of 3D point clouds has been attempted in some previous studies \cite{moosmann2009segmentation,landrieu2017comparison} primarily for segmentation of ground and objects, these studies do not provide a rigorous formulation of normal estimation using the graphical approaches. On the other hand, the normal estimation from multi-model fitting has been attempted under an optimization setup \cite{amayo2018geometric} which can be shown to be a specific instance of our proposed approach. 

In order to develop a fast and robust extraction algorithm, we analyze point clouds through a variant of the proximity graphs, the k-nearest neighbor graph (k-NNG) \cite{toussaint1989some}. The plane extraction problem can then be formulated as a problem of finding the latent structure in the graph. This stems from an underlying assumption that the data samples can be aggregated on a low-dimensional manifold and this manifold can be represented by its discrete conjugate, the graph \cite{rubin2020manifold}. Thus, instead of finding a handful of planar surfaces by segmentation as done by most of the other approaches, we find the normal of each point in point clouds conditioned on the graph. Inspired by analogous research in mining low-dimensional data from high-dimensional data using robust principal component analysis (rPCA) \cite{shahid2016fast}, we impose a graph smoothness constraint on the samples. 

We further extend the proposed formulation to a weighted version to improve the estimation accuracy of the normals at the points near the boundary between two distinct planar surfaces. In the weighted version, we estimate the normals by formulating an optimization problem with a weighted loss function, and update the weights iteratively. Different weighting strategies are chosen intuitively - the dot product of the estimated normals between neighboring points; the inverse of distance between the neighboring points and the product of dot product between the normals and inverse distance. The weights provide a discriminative feature where the preference is given to closer points with similar normal vectors. 

The rest of paper is organized as follows. In Section~\ref{s:related}, we will introduce the related work. In Section~\ref{s:method}, the proposed method is elaborated for normal estimation. Section~\ref{s:result} demonstrates the effectiveness of the method by using a small simulated dataset composed of points sampled from three orthogonal planar surfaces, as well as a large scale synthetic plane estimation benchmarking dataset, SynPeb \cite{schaefer19icra}. A comparison between the proposed method and other benchmark methods are also provided. The paper then ends with a conclusion in Section~\ref{s:discussion}.

\section{RELATED WORK}\label{s:related}
Normal estimation for unorganized point clouds can be broadly classified into regression-based, optimization-based, and Voronoi-based methods. 

\subsection{Regression- and optimization-based methods}
Regression-based normal estimation was first proposed by \cite{hoppe1992surface} where the authors proposed estimating the normal of every individual point in the point cloud as the normal of the local neighborhood of k-neighbors with the assumption that the surface is smooth everywhere (i.e. there are no sudden abrupt normal changes). The normal estimation was performed by fitting a least squares plane via principal component analysis (PCA) which is analogous to minimizing the $L_2$ norm. Given the ease of use and speed of computation of the method, it has been ubiquitously used to estimate the normals of a surface. 

However, there are two main disadvantages of this method: it is very sensitive to noise in the data and it fails to preserve edges since it tends to average the normals of two adjoining surfaces. Various modifications have been suggested to make the surface fitting robust to noise: \cite{fleishman2005robust} replaced PCA with Least Median of Squares (LMS), \cite{lipman2007parameterization} replaced $L_2$ norm with $L_1$ norm which has been theoretically proven to be robust to outliers. Other researchers used weighting to heighten or suppress contribution of neighboring points which has an effect of removing outliers/noise as well such as \cite{pauly2002efficient} who used weighting to give more importance to nearer neighbors. 

Modification to the second disadvantage has been much less studied. \cite{fleishman2005robust} divided the neighborhood into piece-wise planar surfaces by first applying LMS by selecting points lower than a certain threshold in the neighborhood and then running moving least squares (MLS). The drawback of this method was that LMS was sensitive to noise which made the overall method unstable. Recently, \cite{sanchez2020robust} used a well-known robust m-estimator, Geman-McClure estimator, to solve the issue of noise and also subsequently, perform two independent initializations leading to two different solutions to counter the anisotropy around the edge features. However,  m-estimator is known to be very sensitive to initial estimate, a problem common to all nonlinear regression procedures \cite{myers1990classical}, it is sensitive to the scale chosen and finally, convergence with m-estimator is not guaranteed. On the other hand, probabilistic plane extraction (PPE) \cite{schaefer19icra} explores point association to a finite set of planes found by agglomerative hierarchical clustering. There are two drawbacks which can affect the accuracy of the plane extraction - the clustering results vary wildly based on the kind of linkage chosen and the number of possible planes are fixed a priori.

\subsection{Voronoi-based methods}
An alternate approach for normal estimation is to construct some form of Delaunay triangulation (DT) or its conjugate Voronoi diagram for a given point cloud. This method was first proposed by \cite{amenta1999surface} who defined the normal vector of a point as the line through the point to the furthest vertex of the corresponding Voronoi cell, known as polar ball. While this method yielded normals in the noise-free case, in presence of noise it completely broke down. \cite{dey2006normal} expanded this method to deal with noise by finding large polar balls to approximate the normal vector. However, these methods are still not robust to noise in real life LiDAR datasets \cite{sanchez2020robust}.

\section{GRAPH-BASED NORMAL ESTIMATION}\label{s:method}
\subsection{Problem formulation}
The space of measurement data is denoted by $\mathcal{X}\subset \mathbb{R}^3$. Let $\boldsymbol{X}=\left\{\boldsymbol{x}_1,...,\boldsymbol{x}_m\right\}$ denote the set of $m$ measurement points in the LiDAR point cloud, where $\boldsymbol{x}_i\in\mathcal{X}$ is the $i$-th point in $\boldsymbol{X}$. Throughout the paper, we denote scalars by lowercase letters, vectors by lowercase boldface letters, and matrices, sets of vectors by uppercase boldface letters. Let $\boldsymbol{n}_i$ denote the target normal vector at point $\boldsymbol{x}_i$. Our objective is to estimate $\left\{\boldsymbol{n}_i;i=1,\dots,m\right\}$ based on the point cloud data $\boldsymbol{X}$. 

According to the definition of the normal vector \cite{morvan2004approximation}, we would like $\boldsymbol{n}_i$ to be perpendicular to the tangent plane at $\boldsymbol{x}_i$, or equivalently, to be perpendicular to any vector on the tangent plane at $\boldsymbol{x}_i$. Since the tangent plane is unknown, as an empirical approach, we require $\boldsymbol{n}_i$ to be perpendicular to the vectors between $\boldsymbol{x}_i$ and its $k$-nearest neighboring points. In this way, the estimation of $\boldsymbol{n}_i$ is converted to the following optimization problem:
\begin{equation}
    \boldsymbol{\hat n}_i=arg\,\min_{\left\|\boldsymbol{n}\right\|=1} \sum_{\boldsymbol{x}\in\widetilde{\boldsymbol{X}}_i} \langle \boldsymbol{n},\boldsymbol{x}-\boldsymbol{x}_i \rangle^2,\text{ for }i=1,\dots,m,
    \label{eq:separate}
\end{equation}
where $\widetilde{\boldsymbol{X}}_i$ represents the set of $k$-nearest neighboring points of $\boldsymbol{x}_i$, $\left\|\cdot\right\|$ represents the $l^2$ norm, and $\langle\cdot,\cdot\rangle$ represents the inner product.

However, the point cloud data $\boldsymbol{X}$ is often subject to random measurement noise. In this case, $\left\{\boldsymbol{x}-\boldsymbol{x}_i;x\in\widetilde{\boldsymbol{X}}_i\right\}$ may not well approximate the vectors on the tangent plane at $\boldsymbol{x}_i$, and the resultant $\boldsymbol{\hat n}_i$ in Eq.(\ref{eq:separate}) is biased. To improve the robustness of $\boldsymbol{\hat n}_i$, we impose a regularization term to the optimization problem in Eq.(\ref{eq:separate}). Let $\boldsymbol{N}$ be the matrix whose row vectors are $\boldsymbol{n}_i$'s, let $\boldsymbol{n}_{(j)}$ denote the $j$-th column vector of $\boldsymbol{N}$. The regularized optimization problem is then expressed as follows:
\begin{eqnarray}
    \boldsymbol{\widehat N}&=&arg\,\min_{\left\|\boldsymbol{n_i}\right\|=1;i=1,\dots,m} \sum_{i=1}^m \sum_{\boldsymbol{x}\in\widetilde{\boldsymbol{X}}_i} \langle \boldsymbol{n_i},\boldsymbol{x}-\boldsymbol{x}_i \rangle^2,\nonumber\\
    &+&\lambda\sum_{j=1}^3\boldsymbol{n_{(j)}}^\text{T}\boldsymbol{L}\boldsymbol{n_{(j)}},
    \label{eq:penalty1}
\end{eqnarray}
where $\boldsymbol{L}$ is the Laplacian that measures the spatial similarity among the points $i=1,\dots,m$, and $\lambda$ is the tuning parameter. The second term on the right hand side of Eq.(\ref{eq:penalty1}) forces the resultant normal $\boldsymbol{n}_i$ and $\boldsymbol{n}_j$ close to each other when the distance between the points $\boldsymbol{x}_i$ and $\boldsymbol{x}_j$ is small, and hence guarantees the robustness of $\widehat{\boldsymbol{N}}$ with respect to the measurement noise in $\boldsymbol{X}$.

Now we aim to develop the matrix form of the optimization problem in Eq.(\ref{eq:penalty1}) and show that the target problem is a constrained convex optimization problem. Let $\boldsymbol{n}$ denote the vectorized $\boldsymbol{N}$, that is, $\boldsymbol{n}=\left[\boldsymbol{n}_1^\text{T},\dots,\boldsymbol{n}_m^\text{T}\right]^\text{T}$. Let $\boldsymbol{R}_i$ denote the $3 \times 3m$ matrix that projects $\boldsymbol{n}$ to its $i$-th row vector $\boldsymbol{n}_i$. Here the $(3i-2,1)$, $(3i-1,2)$ and $(3i,3)$-th elements in $\boldsymbol{R}_i$ are $1$, while the rest elements are $0$. It can be easily validated that 
\begin{equation}
    \boldsymbol{R}_i \boldsymbol{n}=\boldsymbol{n}_i.
    \label{eq:row}
\end{equation}
Similarly, let $\boldsymbol{C}_j$ be the $m \times 3m$ matrix whose $(1, j)$, $(2, 3+j)$, $\cdots$, $(m, 3m-3+j)$-th elements are $1$, and the rest elements are $0$. Under this definition, $\boldsymbol{C}_j$ projects $\boldsymbol{n}$ to its $j$-th column vector as:
\begin{equation}
    \boldsymbol{C}_j \boldsymbol{n}=\boldsymbol{n}_{(j)}.
    \label{eq:column}
\end{equation}
Let $\mathcal{L}$ denote the loss function in Eq.(\ref{eq:penalty1}). Substituting Eq.(\ref{eq:row}) and Eq.(\ref{eq:column}) into Eq.(\ref{eq:penalty1}) yields:
\begin{eqnarray}
    \mathcal{L}&=&\sum_{i=1}^m \sum_{\boldsymbol{x}\in\widetilde{\boldsymbol{X}}_i} \langle \boldsymbol{R}_i\boldsymbol{n},\boldsymbol{x}-\boldsymbol{x}_i \rangle^2+\lambda\sum_{j=1}^3\boldsymbol{n}^\text{T}\boldsymbol{C}_j^\text{T}\boldsymbol{L}\boldsymbol{C}_j\boldsymbol{n},\nonumber\\
    &=&\sum_{i=1}^m\boldsymbol{n}^\text{T}\boldsymbol{R}_i^\text{T}\boldsymbol{X_{c,i}}^\text{T}\boldsymbol{X_{c,i}}\boldsymbol{R}_i\boldsymbol{n}+\lambda\sum_{j=1}^3\boldsymbol{n}^\text{T}\boldsymbol{C}_j^\text{T}\boldsymbol{L}\boldsymbol{C}_j\boldsymbol{n},\nonumber\\
    &=&\boldsymbol{n}^\text{T}\sum_{i=1}^m\left\{\boldsymbol{R}_i^\text{T}\boldsymbol{X_{c,i}}^\text{T}\boldsymbol{X_{c,i}}\boldsymbol{R}_i\right\}\boldsymbol{n}\nonumber\\
    &+&\lambda\boldsymbol{n}^\text{T}\sum_{j=1}^3\left\{\boldsymbol{C}_j^\text{T}\boldsymbol{L}\boldsymbol{C}_j\right\}\boldsymbol{n},
    \label{eq:penalty2}
\end{eqnarray}
where $\boldsymbol{X}_{c,i}$ is the matrix whose $j$-th row is the difference between the $j$-th point in $\widetilde{\boldsymbol{X}}_i$ and $\boldsymbol{x}_i$. To this end, the loss function is expressed as the summation of two quadratic forms, implying that
\begin{equation}
    \boldsymbol{\hat n}=arg\,\min_{\left\|\boldsymbol{R}_i\boldsymbol{n}\right\|=1;i=1,\dots,m}\mathcal{L}
\end{equation}
is a constrained convex optimization problem with respect to $\boldsymbol{n}$. The optimization problem is well-defined once the Laplacian graph term $\boldsymbol{L}$ is defined. In the next subsection, we will discuss the choice of $\boldsymbol{L}$.

\subsection{Choice of graph Laplacian}
The first hurdle is to choose a graph Laplacian to adequately represent these points. Since the 1980s, there has been a lot of work done in characterizing the distance in irregular point sets (planar and high-dimensional) which uses the notion of proximity graphs. Proximity graphs are geometric graphs in which any two vertices $p, q$ are connected via an edge $e_{pq}$ if and only if there exists a certain exclusion region where no other points are located \cite{mitchell2017proximity}. In this research, we utilize the most basic form of proximity graph, k-nearest neighbor graph (k-NNG) to represent the sampled points. k-NNG is a graph where any two vertices $p, q$ are connected if the distance between $p$ and $q$ is among the $k$-th smallest distances for either $p$ or $q$. 

For a given graph $\boldsymbol{G}$, an adjacency matrix can be constructed which encodes the weights of the edges. We construct the adjacency graph as
\begin{equation*}
	A_{ij} = \begin{cases}
		exp \bigg(-\dfrac{\left\|(\boldsymbol{x}_i - \boldsymbol{x}_j)\right\|^2}{\sigma^2}\bigg) & \text{if $\boldsymbol{x}_j$ is connected to $\boldsymbol{x}_i$}\\
		0 & \text{else}
	\end{cases}
\end{equation*}
with $\sigma = 1$. As previously mentioned, we utilize graph regularization, with a graph encoding the connections to neighboring data samples, $\boldsymbol{G}$. Therefore, we construct an adjacency matrix, $\boldsymbol{A}$ and using the adjacency matrix, we construct the normalized graph Laplacian matrix $\boldsymbol{L}$ as $\boldsymbol{L} = \boldsymbol{I} - \boldsymbol{D}^{-1/2}\boldsymbol{A}\boldsymbol{D}^{-1/2}$ where $\boldsymbol{D}$ is the degree matrix given as a diagonal matrix of the row-wise sum of the adjacency matrix. Given the graph Laplacian, we can solve the optimization problem.

\subsection{Solving optimization using conjugate gradient}
With the Laplacian graph $\boldsymbol{L}$ well defined, the next step is to solve the constrained convex optimization problem in Eq.(\ref{eq:penalty2}). Although $\mathcal{L}$ is the summation of two quadratic forms, the complex equality constraint
\begin{equation}
    \left\|\boldsymbol{R}_i\boldsymbol{n}\right\|=1;i=1,\dots,m,
\end{equation}
makes the optimal solution intractable. Here we adopt the Projected Conjugate Gradient Descent algorithm to obtain the numerical solution. The algorithm iterates two sub-steps until convergence. In the first sub-step, the solution $\boldsymbol{n}$ is updated based on the gradient of $\mathcal{L}$. Specifically, in iteration $t+1$, the normal is updated as
\begin{equation}
    \boldsymbol{n}_{t+1}=\boldsymbol{n}_t+\alpha \nabla_{\boldsymbol{n}}\mathcal{L},
\end{equation}
where
\begin{equation}
    \nabla_{\boldsymbol{n}}\mathcal{L}=2\left(\sum_{i=1}^m\boldsymbol{R}_i^\text{T}\boldsymbol{X_{c,i}}^\text{T}\boldsymbol{X_{c,i}}\boldsymbol{R}_i+\lambda\sum_{j=1}^3\boldsymbol{C}_j^\text{T}\boldsymbol{L}\boldsymbol{C}_j\right)\boldsymbol{n}_t.
\end{equation}

In the second sub-step, $\boldsymbol{n}_{t+1}$ is projected to the space
\begin{equation}
    \left\{\boldsymbol{n}:\left\|\boldsymbol{R}_i\boldsymbol{n}\right\|=1;i=1,\dots,m\right\}
\end{equation}
by normalizing each of its rows to a unit vector. Since the optimization problem is convex, the Projected Conjugate Gradient Descent algorithm returns the unique solution. Our results as detailed in the next section show that this method yields acceptable results in a small number of iterations. 

\begin{figure*}[h!]
	\includegraphics[width=1.0\textwidth]{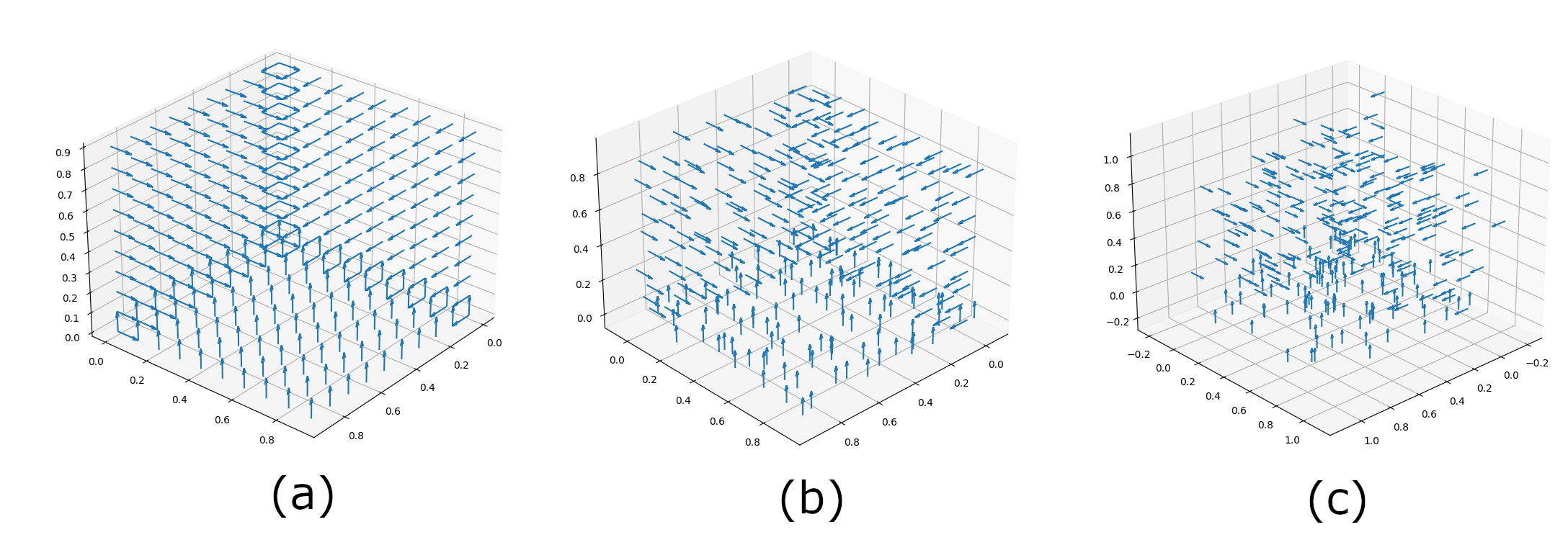}
	\caption{Illustration of the generated point clouds and true normals in the simulation dataset. The tails of the arrows represent the point location, the heads of the arrows show the normal direction. (a) The normals in the ground truth synthetic dataset. (b) The normals with $0.025m$ noise added. (c) The normals with $0.1m$ noise added.}
	\label{f:plot_norm}
\end{figure*}

\subsection{Weighted graph-based normal estimation}
In this subsection, we introduce a weighted version of the graph-based normal estimation algorithm discussed in the previous subsections. The motivation is to correct the estimation bias when $\boldsymbol{x}_i$ is near a boundary of two surfaces that have distinct normals. In this scenario, solving the optimization problem using the loss function in Eq.(\ref{eq:penalty2}) results in a normal lying somewhere in between the normals of the two surfaces, instead of returning the true normal of the surface the point $\boldsymbol{x}_i$ actually belongs to. This is because the loss function requires the estimated normal $\boldsymbol{n}_i$ to be perpendicular to the surface determined by the neighboring point set $\widetilde{\boldsymbol{X}}_i$, which can include points on different surfaces. 

A natural approach to resolve this issue is to place higher weights on the points on the surface $\boldsymbol{S}_i$ where $\boldsymbol{x}_i$ belongs to, and place lower weights on the rest of points in $\widetilde{\boldsymbol{X}}_i$. A natural assumption in this case is that for a point $\boldsymbol{x}_i \in \boldsymbol{S}_i$, where $\boldsymbol{S}_i$ represents the true surface the point is sampled from, majority of the other neighboring points $\widetilde{\boldsymbol{X}}_i$ also belong to $\boldsymbol{S}_i$.
Let $\boldsymbol{W}_i$ denote the target weight matrix, the loss function in Eq.(\ref{eq:penalty2}) can be revised as:
\begin{eqnarray}
    \mathcal{L}&=&\boldsymbol{n}^\text{T}\sum_{i=1}^m\left\{\boldsymbol{R}_i^\text{T}\boldsymbol{X_{c,i}}^\text{T}\boldsymbol{W}_i\boldsymbol{X_{c,i}}\boldsymbol{R}_i\right\}\boldsymbol{n}\nonumber\\
    &+&\lambda\boldsymbol{n}^\text{T}\sum_{j=1}^3\left\{\boldsymbol{C}_j^\text{T}\boldsymbol{L}\boldsymbol{C}_j\right\}\boldsymbol{n}.
    \label{eq:penalty3}
\end{eqnarray}

which can be seen as similar to the weighted PCA. 
In this case, the gradient can be computed as
\begin{equation}
	\nabla_{\boldsymbol{n}}\mathcal{L}=2\left(\sum_{i=1}^m\boldsymbol{R}_i^\text{T}\boldsymbol{X_{c,i}}^\text{T}\boldsymbol{W}_i\boldsymbol{X_{c,i}}\boldsymbol{R}_i+\lambda\sum_{j=1}^3\boldsymbol{C}_j^\text{T}\boldsymbol{L}\boldsymbol{C}_j\right)\boldsymbol{n}_t.
	\label{eq:gradient2}    
\end{equation}

Since the true normal of the points in $\widetilde{\boldsymbol{X}}_i$ is unknown, we would like to refer to the geometric information of the point set $\widetilde{\boldsymbol{X}}_i$ and provide the following three choices of the weight matrix $\boldsymbol{W}_i$. Let $w_{i,jk}$ denote the element in $\boldsymbol{W}_i$ which is corresponding to the pairwise distance between the $j$-th and $k$-th points in $\widetilde{\boldsymbol{X}}_i$.

\begin{itemize}
	\item \textbf{dot product between normals} - weight of each pair of points $j$ and $k$ is given by the dot product computed between the normals as:
	\begin{equation}
		\begin{aligned}
			w_{i,jk} =  \abs{\boldsymbol{n_j} \cdot \boldsymbol{n_k}}
		\end{aligned}
	\end{equation}
	\item \textbf{distance between points} - weight of each pair of points $j$ and $k$ is given by the inverse of Euclidean distance as:
	\begin{equation}
		\begin{aligned}
			w_{i,jk} = \frac{1}{\left\|\boldsymbol{x}_j-\boldsymbol{x}_k\right\|}
		\end{aligned}
	\end{equation}
	
	\item \textbf{combination of dot product and distance} - weight of each pair of points $j$ and $k$ is given by the product of dot product and inverse distance as:
	\begin{equation}
		\begin{aligned}
			w_{i,jk} = \frac{\boldsymbol{n_j} \cdot \boldsymbol{n_k}}{\left\|\boldsymbol{x}_j-\boldsymbol{x}_k\right\|}
		\end{aligned}
	\end{equation}

\end{itemize}

We provide the results in the following section for each of the weighting strategies and compare against the other competing approaches. As a summary, we present the proposed algorithm as follows.

\begin{algorithm}
	\SetAlgoLined
	\SetKwInOut{Input}{input}
	\SetKwInOut{Output}{output}
	
	\Input{a set of point cloud data $\boldsymbol{X}$}
	
	\Init{}{neighboring point sets $\boldsymbol{\widetilde{{X}}}_i$, $i=1,\dots,m$\\
		graph Laplacian $\boldsymbol{L}$\\
		projection matrices $\boldsymbol{R}_i,\boldsymbol{C}_i$, $i=1,\dots,m$\\
		centered matrix $\boldsymbol{X}_{c,i}$, $i=1,\dots,m$\\
		tuning parameter $\lambda$, initial normal $\boldsymbol{n}_0$, iteration $t=0$, learning rate $\alpha$\\}
	
	\While{$\left\|\boldsymbol{n}_t-\boldsymbol{n}_{t-1}\right\|\geq \varepsilon$}{
		Evaluate $\boldsymbol{W}_i$ based on $\boldsymbol{n}_t$ for $i=1,\dots,m$\\
		Evaluate $\mathcal{L}$ in Eq.(\ref{eq:penalty3})\\
		Evaluate $\nabla_{\boldsymbol{n}}\mathcal{L}$ in Eq.(\ref{eq:gradient2})\\
		$\boldsymbol{n}_{t+1}\gets\boldsymbol{n}_t+\alpha \nabla_{\boldsymbol{n}}\mathcal{L}$\\
		$t\gets t+1$\\
	}
	
	\Output{the estimated normal $\hat{\boldsymbol{n}}=\boldsymbol{n}_t$}
	\caption{Graph-based Normal Estimation}
	\label{a:overall}
\end{algorithm}

\begin{figure*}[h!]
	\includegraphics[width=1.0\textwidth]{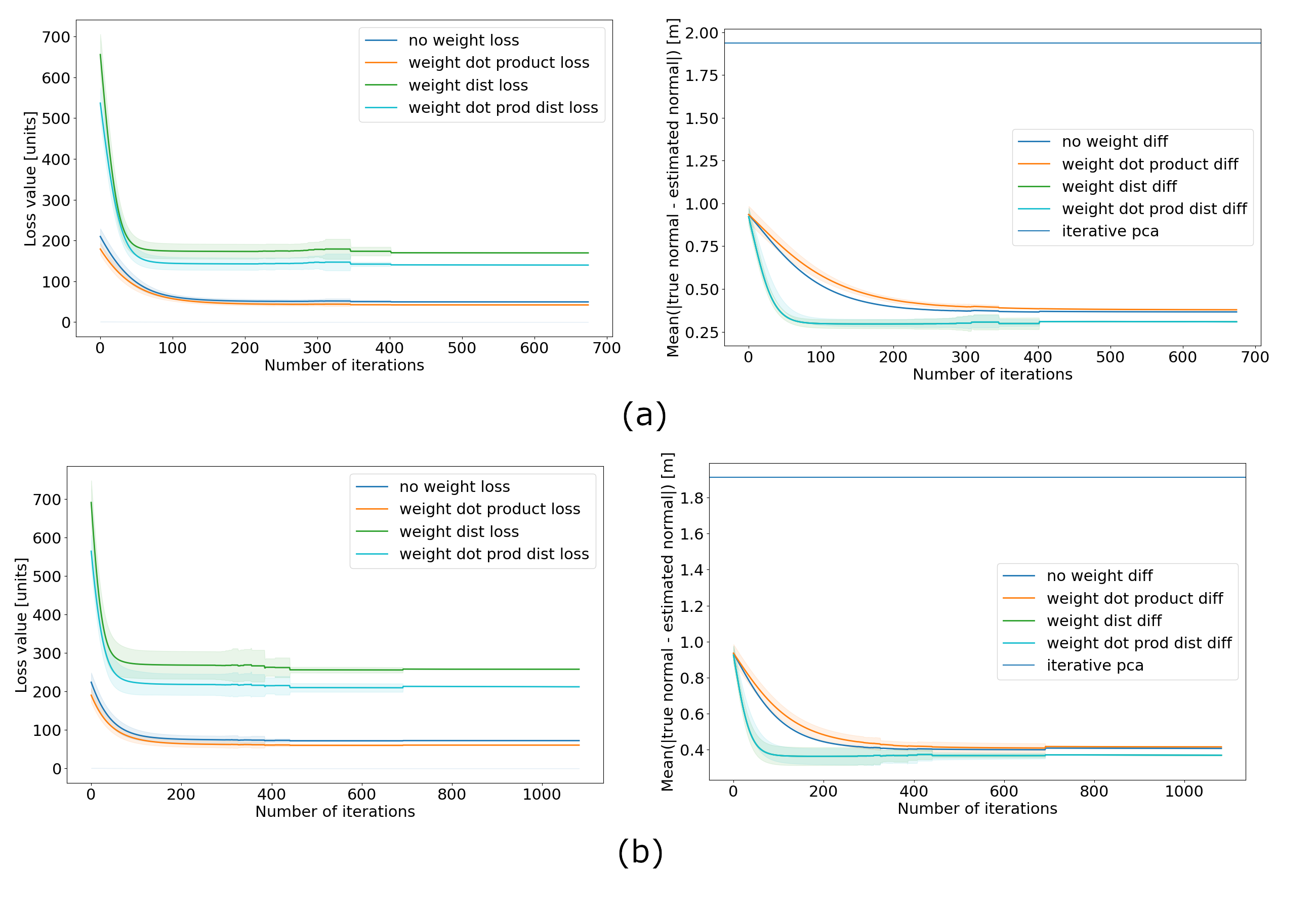}
	\caption{The mean performance under different weight strategies is depicted by the solid lines. The shaded area illustrates the corresponding confidence interval. Comparison of the performance of the proposed algorithm under different weighting strategies. Left panel: loss function; right panel: mean squared bias from the ground truth normals. (a) Results for random noise with a standard deviation of $0.025m$; (b) results for random noise with a standard deviation of $0.05m$.}
	\label{f:plot_loss_diff_1}
\end{figure*}

In regard to the computational complexity, the computing time depends on the number of iterations, the evaluation of the loss function and its gradient in Eq.(\ref{eq:penalty3}) and Eq.(\ref{eq:gradient2}). In fact, the computational time of Eq.(\ref{eq:penalty3}) and Eq.(\ref{eq:gradient2}) is dominated by the matrix multiplication of maximal size $3k\times 3k$ and the summation over all the data points, which is $O\left(m\times k^3\right)$. Let $T$ denote the total number of iterations, the overall computational complexity of the proposed method is hence $O\left(m\times k^3\right)$. This indicates that the computational time scales up linearly with respect to the number of points and iterations, which can be applied to high-volume point clouds. On the other hand, a large number of neighbors $k$ dramatically increases the computational load, while improving the estimation accuracy for a non-smooth surface with a high measurement noise level. Given the trade-off between the estimation accuracy and computational time, it is recommended to determine the number of neighbors based on the prior information on the surface smoothness and measurement noise level.

\section{RESULTS}\label{s:result}
\subsection{Simulation setup}\label{ss:simulation}
We will use a toy example in subsection~\ref{ss:simulation} to show the effectiveness of the proposed method in extracting normals from a small yet noisy dataset. In subsection~\ref{ss:case}, we will use the SynPeb dataset with $250,000$ points to demonstrate the method's applicability to high-volume complicated real data \cite{schaefer19icra}. The combination of the two studies validates the effectiveness of the proposed method under different scenarios.

We introduce the simulation setup in this subsection in which the efficacy and robustness of our proposed approach can be tested and compared to other state-of-the-art methods. We utilize three orthogonal planar surfaces representing the 3D coordinates and sample 100 points from each surface in a grid pattern as shown in Fig. \ref{f:plot_norm}(a). The 3D locations of these points are then disturbed via adding the random Gaussian noise with a zero mean and standard deviations at four different levels - $\{0.025, 0.05, 0.075, 0.1\}$. The smallest noise level corresponds to the highest signal-to-noise (SNR) ratio, surveying grade LiDAR noise while the highest noise level corresponds to depth cameras such as Kinect as shown in Fig. \ref{f:plot_norm}(b) and \ref{f:plot_norm}(c). 

\begin{figure*}[h]
	\includegraphics[width=1.0\textwidth]{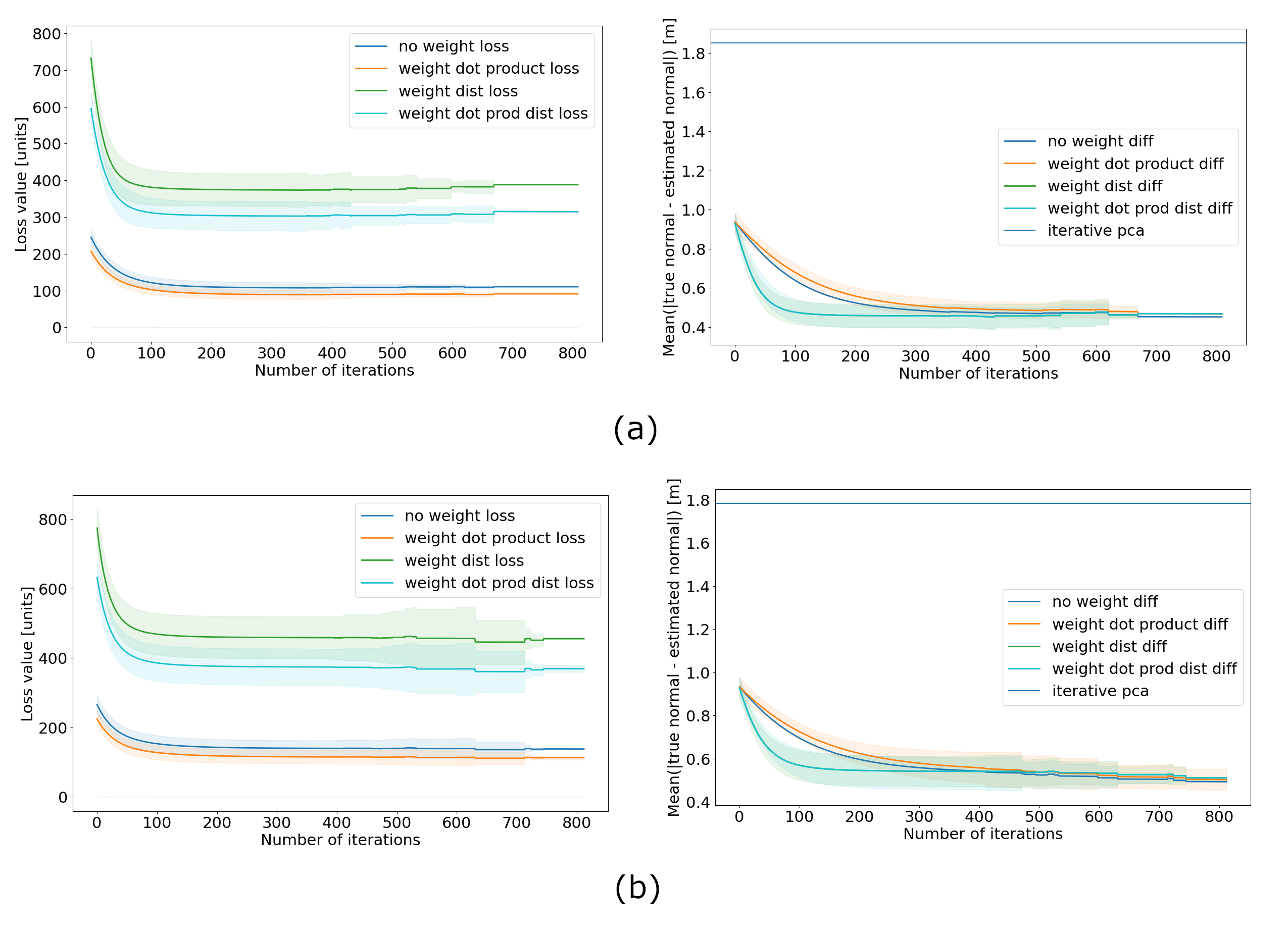}
	\caption{Plotting the loss of the proposed algorithm under different weighting strategies in left side of figure and mean difference between the true and estimated normals in right side of figure. (a) gives the results for 0.075 m random noise added and (b) gives the results for 0.1 m random noise added}
	\label{f:plot_loss_diff_2}
\end{figure*}

The simulation study is conducted $30$ iterations at each noise level. In each iteration, we generate the dataset, apply the proposed approach to the datasets, and check the estimation performance. To illustrate the estimation uncertainty, we plot both the mean and $\pm3$ standard deviation (corresponding to the $99\%$ confidence intervals) over the $30$ iterations. We showcase the training loss in Eq.(\ref{eq:penalty3}) (in the left panels of Fig. \ref{f:plot_loss_diff_1} and Fig. \ref{f:plot_loss_diff_2}) and the estimation bias of the normals from the ground truth, $\left\|\boldsymbol{\hat{n}}-\boldsymbol{n}\right\|$ (in the right panels of Fig. \ref{f:plot_loss_diff_1} and Fig. \ref{f:plot_loss_diff_2}). Particularly, we compare the proposed method under the basic optimization setup (referred in the figures as \textit{no weight}) and under the three weighting strategies (referred in the figures as \textit{weight dot prod}, \textit{weight dist} and \textit{weight dot prod dist} respectively). We compare our proposed approach with our reproduction of the robust normal estimation proposed by \cite{sanchez2020robust} which is illustrated in a horizontal line in the difference plots. The left panels of Fig. \ref{f:plot_loss_diff_1} and Fig. \ref{f:plot_loss_diff_2} indicate that the proposed algorithm consistently converges under different choices of weighting strategies. Among them, weighting using a combination of dot product and distance shows a higher convergence speed than the other weighting strategies. The right panels of Fig. \ref{f:plot_loss_diff_1} and Fig. \ref{f:plot_loss_diff_2} imply that the weights using a combination of dot product and distance achieves the minimal bias comparing to the other weighting strategies. It is worth noting that the formulation without weighting results in an estimation bias of $1.90$ since the normals near the boundaries between two different surfaces are not correctly estimated. With the weighting strategy using a combination of dot product and distance introduced, the estimation bias is reduced to about $0.30$, demonstrating the necessity of the iterative weighting procedure.

\begin{figure*}[h]
	\includegraphics[width=1.0\textwidth]{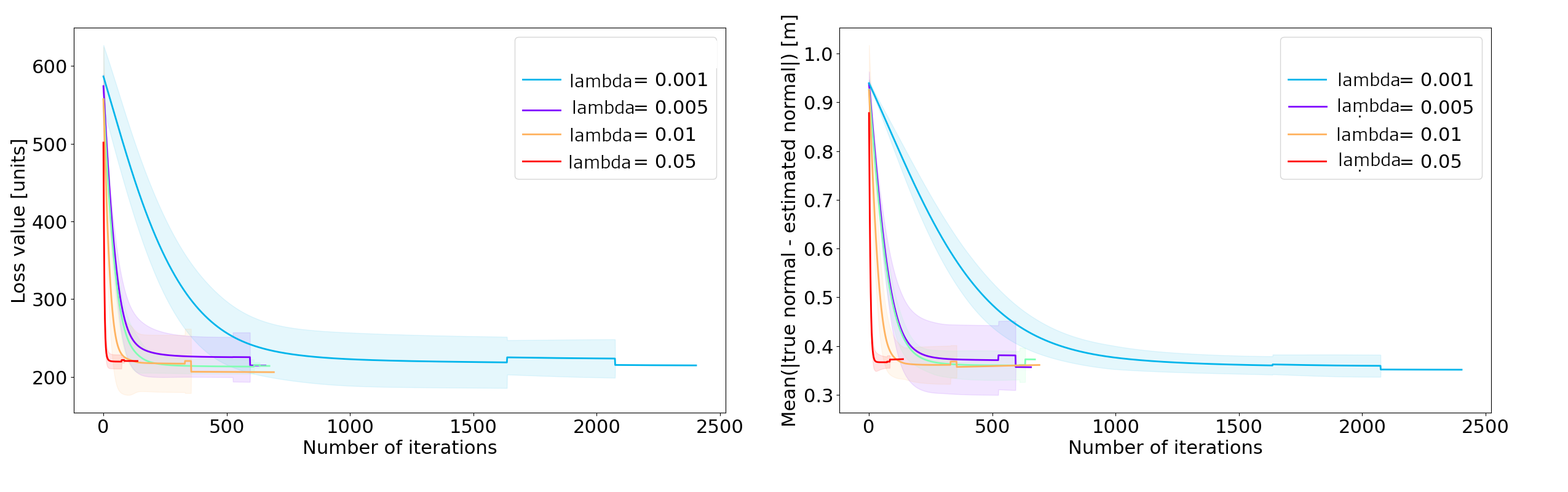}
	\caption{Plotting the loss of the proposed algorithm under different  tuning parameter values in left side of figure and mean difference between the true and estimated normals in right side of figure. In all of the cases, we sample 10 different randomly chosen datasets with 0.05m of noise.}
	\label{f:plot_loss_diff_alpha}
\end{figure*}

For all the experiments conducted so far, we utilize a constant tuning parameter $\lambda = 0.01$. However, in other learning literature (for e.g. \cite{do2007efficient}), it has been mentioned that the value of the penalty is more important than the choice of the regularization. In order to understand the effect of the hyperparameter, we perform a grid search for the  tuning parameter $\lambda$ over the set $\left\{0.001, 0.005, 0.01, 0.05\right\}$ and showcase the loss values and estimation bias over all iterations for 10 randomly corrupted datasets at each level. The results for the loss and bias using the third weighting strategy for the different tuning parameters at all iterations are given in Fig. \ref{f:plot_loss_diff_alpha}. The figure suggests to use a $\lambda$ greater than $0.005$ to achieve a low estimation bias. The figure also implies that a higher tuning parameter increases the convergence speed with the highest $\lambda$ value leading to the fastest convergence. This is natural because a higher $\lambda$ sets more regularization on the similarity of the estimated normals in neighboring points, which reduces the efforts in searching for an individual normal for each data point.

\subsection{SynPeb dataset results}\label{ss:case}
The SynPeb dataset was released by \cite{schaefer19icra} as a comprehensive benchmarking dataset for planar estimation from depth images which got rid of some of the glaring problems of the original planar extraction benchmarking SegComp Perceptron Dataset \cite{hoover1996experimental}, namely, errors in noise and labeling. Also, SynPeb can provide accurate normal information on account of it being a simulated dataset while being very complex. Although \cite{hoover1996experimental} presented a cluster-based normal estimation method and showed a good estimation performance, the estimation procedure highly depends on the quality of clustering, which may not always be high in real applications.

Alternatively, our proposed formulation does not require any prior information on clustering. Given the 3D point clouds, we first apply the proposed method to estimate the normals, compute the pairwise dot products among all the estimated normals, and then directly apply any graph-based clustering method \cite{schaeffer2007graph} by using the pairwise dot product as the similarity metric between a pair of points. We set the threshold as $0.95$ and consider a pair of points come from the same cluster when their corresponding pairwise dot product exceeds the threshold. We would like to point out that there might be additional noise based on the clustering algorithm chosen for this task but we report the entire error as part of normal estimation (assuming that the clustering algorithm is perfect).

Table \ref{tab:synpeb} provides the results of the SynPeb dataset by the proposed approach and compares it to the state-of-the-art methods that use the same dataset for plane extraction. The table indicates that our proposed approach outperforms the state-of-the-art methods in terms of correct number of points extracted per plane as well as the root mean square error, including the best state-of-the-art method, PPE. On the other hand, the graph Laplacian regularization avoids the excessively large number of segmentations induced by neighboring normals not being similar. Compared to PPE, our proposed method does not constrain the number of planes and does not start with the clustering which can introduce additional error. Of course, as with PPE, the increased accuracy comes at the cost of increased computational speed and spatial complexity.

\begin{table*}[h!]
    \centering
    \begin{tabular}{c c c c c c c c c}
     Method & fraction [\%]  & correct [\%] & RMSE [mm] & $\alpha$ [$\degree$] & $n_o$ & $n_u$ & $n_m$ & $n_s$\\
     PEAC \cite{feng2014fast} & 29.1 & 60.4 & 28.6 & – & 0.7 & 1.0 & 26.7 & 7.4\\
     MSAC \cite{torr2000mlesac} & 7.3 & 35.6 & 34.3 & – & 0.3 & 1.0 & 36.3 & 10.9\\
     PPE \cite{schaefer19icra} & 73.6 & 77.9 & 14.5 & – & 1.5 & 1.1 & 7.1 & 16.5\\
     Graph-based normal (proposed)   & 78.5 & 81.3 & 14.1 & 2.1 & 1.3 & 1.1 & 8.2 & 19.1
    \end{tabular}
    \caption{Results from plane extraction algorithms from SynPeb dataset. The results for the first three methods are taken from Schaefer et al., 2019. Among the header variables, $fraction$ indicates the fraction of points correctly labeled; $correct$ represents the percentage of correctly labeled points in each plane; $RMSE$ represents how close the extracted planes represent the point cloud; $\alpha$ provides the mean angular deviation; $n_o$, $n_u$, $n_m$, and $n_s$ represent the number of oversegmented, undersegmented, missing, and spurious planes compared to the ground-truth segmentation.}
    \label{tab:synpeb}
\end{table*}

\section{DISCUSSIONS AND CONCLUSIONS}\label{s:discussion}

We propose a method for estimating normals from LiDAR datasets using a graph-based formulation. We show that the graph-based formulation provides a natural way of extracting normals while being robust to measurement noise. The weighted version of the optimization algorithm also naturally takes the edge preservation into account by giving more preference to points in the neighborhood either with similar normals or those which are closer or a combination of both. 

Our proposed method is truly a combination of the regression based methods and Voronoi based methods. We solve a single nonlinear optimization taking the neighborhood of each individual point in the point cloud as a locally planar surface similar to the PCA-based approaches. However, we also add a penalization based on the graph Laplacian of a k-NNG. Delaunay Triangulation, as used in the Voronoi based approaches, is a special case of a proximity graphs group of which k-NNG is also a member. Also, the graph Laplacian can be seen as a measure of the divergence of the gradient which is connected to the polar ball introduced in the second set of methods. 

As mentioned previously, our proposed approach also generalizes multi-model planar fitting given by \cite{amayo2018geometric}. Their loss function is given as:
\begin{equation}
    \begin{centering}
    \sum_{l=1}^L \Big(\int_\Omega \rho_l(\mathbf{u}, \phi_l(\mathbf{u}))  + \lambda \omega_\mathcal{N} R(\nabla_\mathcal{N} \phi_l(\mathbf{u}))d\Omega \Big) + \beta L
    \end{centering}
\end{equation}
where the first term represents the cost of a point supporting a particular model with $\phi_l(\mathbf{u})$ representing an indicator function for a point being part of a model. The second term promotes homogeneity with $R$ being the penalty function for neighboring points not belonging to the same model. Thus, the second term can be seen as a specific instance of the k-NNG where only the immediate neighborhood is considered. 

In this work, only k-NNG graph is explored. But the topic of RNG is very rich with different kinds of proximity graphs promoting different strengths of neighboring connections. Our aim in future work is to compare and contrast different proximity graphs. Another future work would be to utilize graph theoretical methods for registration of point clouds. We believe that this work would promote a larger focus into use of graphs in LiDAR data.

{
	\begin{spacing}{1.17}
		\normalsize
		\bibliography{isprs_ref} 
	\end{spacing}
}

\end{document}